\documentclass[pdflatex,sn-mathphys-num]{sn-jnl}

\usepackage{graphicx}%
\usepackage{multirow}%
\usepackage{amsmath,amssymb,amsfonts}%
\usepackage{amsthm}%
\usepackage{mathrsfs}%
\usepackage[title]{appendix}%
\usepackage{xcolor}%
\usepackage{textcomp}%
\usepackage{manyfoot}%
\usepackage{booktabs}%
\usepackage{algorithm}%
\usepackage{algorithmicx}%
\usepackage{algpseudocode}%
\usepackage{listings}%

\usepackage{acronym}
\usepackage{graphicx}
\usepackage{subcaption} 
\usepackage{tikz}

\acrodef{CCE}{colon capsule endoscopy}
\acrodef{CNN}{convolutional neural network}
\acrodef{PCA}{principal component analysis}
\acrodef{CAM}{class activation map}

\theoremstyle{thmstyleone}%
\theoremstyle{thmstyletwo}%

\theoremstyle{thmstylethree}%

\begin{document}

\title[Using deep learning for predicting cleansing quality of colon capsule endoscopy images]{Using deep learning for predicting cleansing quality of colon capsule endoscopy images}


\author*[1]{\fnm{Puneet} \sur{Sharma}}\email{puneet.sharma@uit.no}

\author[2]{\fnm{Kristian} \sur{Dalsbø Hindberg}}

\author[3,5]{\fnm{Benedicte} \sur{Schelde-Olesen}}

\author[3,5]{\fnm{Ulrik} \sur{Deding}}

\author[4]{\fnm{Esmaeil } \sur{S. Nadimi}}

\author[4]{\fnm{Jan-Matthias } \sur{Braun}}
\author[]{on behalf of the AICE consortium}

\affil*[1]{\orgdiv{Department of Automation and Process Engineering (IAP), \orgname{UiT The Arctic University of Norway},  \state{Tromsø}, \country{Norway}}}

\affil*[2]{\orgdiv{Department of Physics and Technology (IFT), \orgname{UiT The Arctic University of Norway},  \state{Tromsø}, \country{Norway}}}

\affil[3]{\orgdiv{Department of Clinical Research, , \orgname{University of Southern Denmark, }, \state{Odense}, \country{Denmark}}}

\affil[4]{\orgdiv{The Maersk Mc-Kinney Møller Institute, Faculty of Engineering, \orgname{University of Southern Denmark, }, \state{Odense}, \country{Denmark}}}

\affil[5]{\orgdiv{Department of Surgery, \orgname{ Odense University Hospital}, \state{Odense}, \country{Denmark}}}


\abstract{In this study, we explore the application of deep learning techniques for predicting cleansing quality in \acf{CCE} images. Using a dataset of 500 images labeled by 14 clinicians on the Leighton–Rex scale (Poor, Fair, Good, and Excellent), a ResNet-18 model was trained for classification, leveraging stratified $K$-fold cross validation to ensure robust performance. To optimize the model, structured pruning techniques were applied iteratively, achieving significant sparsity while maintaining high accuracy. Explainability of the pruned model was evaluated using Grad-CAM, Grad-CAM++, Eigen-CAM, Ablation-CAM, and Random-CAM, with the ROAD method employed for consistent evaluation. Our results indicate that for a  pruned model, we can achieve a cross-validation accuracy of 88\% with 79\% sparsity, demonstrating the effectiveness of pruning in improving efficiency from 84\% without compromising performance. We also highlight the challenges of evaluating cleansing quality of \ac{CCE} images, and emphasize the importance of explainability in clinical applications, and the challenges associated with using the the ROAD method for our task. Finally, we employ a variant of adaptive temperature scaling to calibrate the pruned models for an external dataset.}




\maketitle

\section{Introduction}
\label{sec:introduction}

Advancements in deep learning have profoundly reshaped the field of medical imaging, finding applications in areas such as radiology, pathology, cardiology, and beyond~\cite{bioengineering10121435}. Deep learning-driven diagnostic tools enhance the efficiency of analyzing complex medical images, enabling earlier disease detection and ultimately leading to improved patient outcomes~\cite{bioengineering10121435}.

Capsule endoscopy has become a vital tool for gastrointestinal (GI) tract examinations in a minimally invasive manner~\cite{SPADA201519, Hosoe2021}. It enables clinicians to visualize the GI tract in detail, providing critical insights into conditions such as  Crohn's disease, bleeding, inflammation, ulcers, polyps, and tumors~\cite{Mota2024, Koffas2022, pennazio2008capsule, blanes2018capsule, wang2021locally, tashk2019fully, Sahafi2022}. However, its effectiveness is often compromised by low cleansing quality of the captured images~\cite{Mota2024}. Residual fecal matter, mucus, blood, bubbles, and debris can obscure key diagnostic features, making it challenging for clinicians to interpret the images accurately. In other words, a proper assessment of the cleansing is required to guarantee that sufficient mucosa is seen and to classify the investigation as complete~\cite{SPADA201519}. 
 
This issue is further exacerbated by significant inter- and intra-observer variability in cleansing assessment, as highlighted in the study by Buijs et al.~\cite{Bujis2018b, Buijs2018}. It not only impacts diagnostic accuracy, but also increases the time required for manual review, highlighting the need for automated solutions and to standardize and streamline the assessment process. 

Deep learning, particularly through~\acp{CNN}, utilizes multiple processing layers to learn data representations at varying levels of abstraction~\cite{LeCun2015}. These techniques have significantly advanced the state-of-the-art in areas such as speech recognition, visual  object detection, image segmentation~\cite{Tashk2022} and image classification~\cite{LeCun2015}.

While deep learning models are known for their high accuracy, their deployment in clinical settings often face challenges related to computational efficiency and explainability. Large, dense neural networks can be computationally expensive, making them unsuitable for resource-constrained environments such as portable medical devices or closed systems essential for hospital use. Explainability is another critical aspect of deploying deep learning models in clinical applications~\cite{nadimi2024integrationartificialintelligencecolon}, where both clinicians and patients often require transparent and interpretable results to trust and validate the predictions made by artificial intelligence systems~\cite{Ploug2021population}.

In this paper, we explore three key questions: First, is it possible to prune a model without compromising on the accuracy associated with cleansing quality estimation of CCE images? Second, can metrics from CAM-based methods be used to explain the results of the pruned models? Third, can we perform model calibration on the pruned models for an external dataset of CCE images?

\section{Data}
\label{sec:data}

In this section, we discuss the two data sets used for training and testing of the proposed models.

\begin{figure}[!ht]
\centering
\includegraphics[width = 0.65\textwidth]{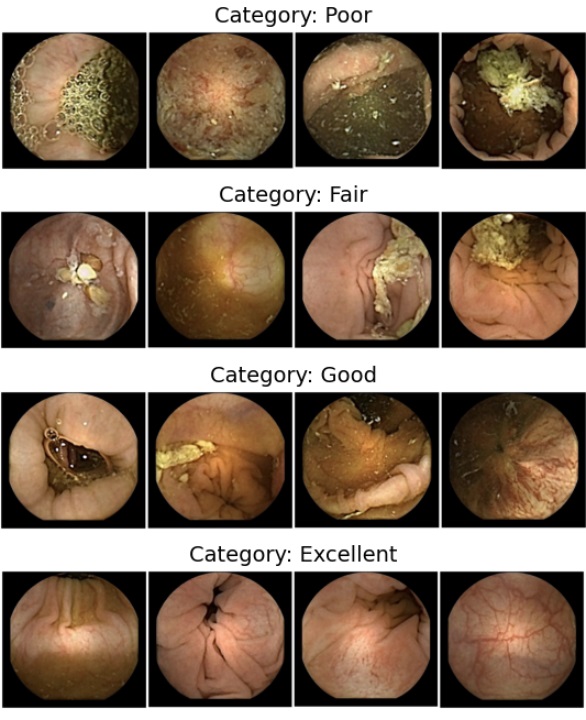}
\caption{Example images of the different Leighton–Rex scale classes}
\label{fig:image_examples}
\end{figure}

The data used for cross-validation in this study is retrieved from the Danish Care For Colon 2015 trial~\cite{Kaalby2020, Baatrup2025}.
The dataset~\cite{ScheldeOlesen2023} consists of 500 images that have been labeled by 14 clinicians (out of a panel of 27) using the Leighton–Rex scale~\cite{tabone2021} as Poor, Fair, Good, and Excellent in cleansing quality. In order to use a ground truth label for each image, we use the mode value of the labels provided by the 14 different clinicians. Figure~\ref{fig:image_examples} shows random examples from the four classes. Assigning a consistent label for each image can be challenging for the different clinicians involved as they may have varying levels of experience, training, or familiarity with specific conditions or nuances in capsule endoscopy images~\cite{Spada2019,ScheldeOlesen2024}. Although cleansing can refer to the absence of fecal matter, mucus, blood, or bubbles, the acceptable threshold for these factors can vary~\cite{ScheldeOlesen2023}. Figure~\ref{fig:image_examples} shows examples of artifacts such as mucus, bubbles, and fecal matter. 

The images were preprocessed using a circular and a border mask to remove the text along the borders. Each image is preprocessed by normalizing it to have a zero mean and unit standard deviation, for training in deep learning.

The data used for testing the proposed models is collected in a separate study where image cleansing quality was rated by 3 different clinicians not related to the previous study.  The image labels where all three clinicians agreed were used as ground truth for testing the performance of the proposed models. This external dataset consists of 396 images, where all the clinicians agree on their cleansing quality.

\section{Theory}
\label{sec:theory}

We have focused on the ResNet-18 (Residual network) architecture~\cite{resnetpaper} in this study. The key feature of ResNets is that it solves the problem of disappearing gradients in deep neural networks by introducing the so-called skip connections~\cite{resnetpaper}. Skip connections allow gradients to flow directly through layers, avoiding the vanishing-gradient problem, which enables training deep networks with hundreds or even thousands of layers without degradation in performance. The ResNet architecture is available in different variants, e.g., ResNet-18, ResNet-34, ResNet-50, ResNet-101, and ResNet-152, where the integer indicates the number of layers.

For training the models used in our experiments, we employed stratified $K$-fold cross validation. $K$-fold cross-validation is used in machine learning to assess how well a model generalizes to unseen data, and hence reduce overfitting~\cite{kfoldvalidpaper}. The dataset is divided into $K$ equal-sized subsets called folds. The model is trained on $K-1$ folds, and the remaining fold is used for validation~\cite{kfoldvalidpaper}. This process is repeated $K$ times, each time using a different fold as the validation set. In Stratified $K$-fold, each fold has a proportional representation of each class, which ensures that all classes are well-represented in both training and validation sets.

Structured pruning is a technique used in deep learning to reduce the complexity of neural networks by removing entire groups of parameters, such as neurons in fully connected layers, channels of layers in \acp{CNN}, rather than individual weights~\cite{Hoefler2021}. This technique improves model efficiency, reduces computational costs, and allows deployment on resource-constrained devices while maintaining acceptable performance. Structured pruning strategies often utilize L1-norm or L2-norm-based methods, which remove channels with the lowest magnitudes under the assumption that low-magnitude channels contribute less to model performance~\cite{He2024}. In addition, pruning strategies include: variance based~\cite{Grunwaldbook, Gao2019, Hoefler2021}, and knowledge-distillation~\cite{Aghli2021,Hoefler2021}. A study~\cite{gale2019statesparsitydeepneural}, highlights that magnitude-based pruning is widely used and achieves results comparable to or even better than more complex methods. This suggests that straightforward pruning strategies can be highly effective for large-scale neural networks.

It is important to note that pruning methods based on an aggressive pruning criterion i.e., removing a large percentage of weights based solely on their magnitude, can cause entire layers to collapse~\cite{liao2024tilllayerscollapsecompressing}, making the network non-functional~\cite{Tanaka2020}. To mitigate this, Tanaka et al.~\cite{Tanaka2020} suggest to prune in multiple steps such that the lowest-ranking weights, i.e., those with the least contribution can be pruned in multiple steps rather than all at once. In addition, we can employ layer-wise local pruning namely pruning the same rate at each layer which ensures that every layer retains some parameters.

\begin{figure}[!ht]
\begin{center}
    \begin{tikzpicture}[node distance=3cm, thick]
    
        \node (train) [draw, rectangle, minimum width=2.5cm, minimum height=1cm] {Train};
        \node (prune) [draw, rectangle, minimum width=2.5cm, minimum height=1cm, right of=train] {Prune};
        \node (fine_tune) [draw, rectangle, minimum width=2.5cm, minimum height=1cm, right of=prune] {Fine-Tune};
        
        \draw[->] (train) -- (prune);
        \draw[->] (prune) -- (fine_tune);
        \draw[dotted, ->] (fine_tune.south) to[out=-90, in=-90] (prune.south);
    \end{tikzpicture}
\end{center}
\caption{The classic three-step framework—train, prune, and fine-tune from the study by~\cite{Han2015}.}
\label{fig:threestepclassic}
\end{figure}
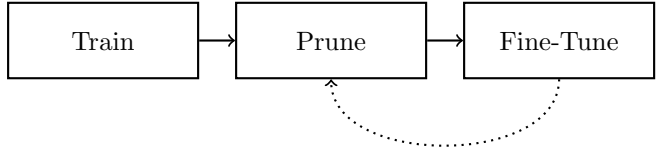

Figure~\ref{fig:threestepclassic} shows a classical three-step framework (train, prune, and fine-tune), as proposed by Han et al.~\cite{Han2015}. In the first step, a dense deep learning model is training on the target dataset. This initial training allows the model to learn the data representations and identify which connections (weights) are significant for accurate predictions. In the second step, pruning is done, where connections deemed less critical are removed to create a sparser network. The authors employed a magnitude-based pruning approach, removing weights with the smallest absolute values. The final fine-tuning step retrains the sparser model to adjust the remaining weights. This step helps the network to recover its accuracy, ensuring that the efficiency gains from pruning do not come at the cost of predictive performance. In multi-step/iterative pruning, steps 2 and 3 are repeated a number of times or until convergence is achieved.

To compare the pruned models regarding explainability, we employ the Remove and Debias (ROAD) framework~\cite{ROAD2022}. In ROAD framework, the explainability of machine learning models is evaluated by applying controlled perturbations to the input data and analyzing the model's response. Specifically, ROAD removes parts of the input (e.g., pixels in an image) and replaces them using a linear combination of neighboring values with added noise to prevent information leakage~\cite{ROAD2022}. This ensures that the perturbed inputs remain realistic while eliminating unintended biases in the evaluation process. According to the authors, unlike traditional methods, ROAD avoids the computationally expensive step of retraining models after each input perturbation, reducing computational costs by up to 99\%. By standardizing this perturbation and replacement process, ROAD enables consistent and unbiased comparisons across different explainability strategies, making it a robust and efficient tool for evaluating model interpretability.

The ROAD method links with faithfulness-based evaluation metrics~\cite{tomsett2020sanity}, which rather than just looking at heatmaps, systematically remove important pixels (found by the post-hoc attribution methods~\cite{Ribeiro2016,Mishra2017}). Where the change in confidence after removing important pixels $S$ is found as
\begin{equation}
S = \sum_{n=1}^{N} f(X-X_{i}) -f(X), 
\label{eq:first}
\end{equation}
where $f(X)$ is the model's confidence score for the original image $X$, $X_{i}$ is image with most important pixel $i$ removed, $f(X-X_{i})$ is the model’s confidence after removing pixel $i$, and $N$ is number of pixels. A higher $S$ can mean a better attribution map, i.e., removing highlighted pixels reduces the model’s confidence significantly, which can indicate that the attribution method correctly identifies important features. On the other hand, a lower $S$ implies that removing highlighted pixels does not affect the confidence score, i.e., the attribution map is not accurate.

For ROAD, equation~(\ref{eq:first}) is modified as 
\begin{equation}
\begin{aligned}
S_{\text{MoRF}, \theta} = \sum_{n=1}^{N_{\theta}} \big[ f(X - X_{\text{MoRF}, i}) - f(X) \big], \\
S_{\text{LeRF}, \theta} = \sum_{n=1}^{N_{\theta}} \big[ f(X - X_{\text{LeRF}, i}) - f(X) \big],
\end{aligned}
\end{equation}
where $X_{\text{MoRF,i}}$ is the image with the $i$-th most important pixel removed (MoRF strategy), $S_{\text{MoRF},\theta}$ is total drop in confidence after removing $\theta \%$ of the most important pixels denoted here by $N_{\theta}$, $X_{LeRF,i}$ is the image with the $i$-th least important pixel removed (LeRF strategy), and $S_{\text{LeRF},\theta}$ is the total drop in confidence after removing $\theta \%$ of the least important pixels.

$S_{MoRF}$ and $S_{LeRF}$ can be combined to create a single ROAD metric $S_{\text{ROAD}}$ that is a combination of LeRF and MoRF and by employing different thresholds as~\cite{jacobgilpytorchcam}

\begin{equation}
S_{\text{ROAD}} =  \sum_{\theta \in \{20, 40, 60, 80\}} \frac{S_{\text{MoRF}, \theta} - S_{\text{LeRF}, \theta}}{2}.
\label{eq:modified}
\end{equation}
where $\theta$ denotes the four different percentile thresholds [20,40,60,80].

In deep learning, model calibration is defined as the process of aligning a model's predicted probabilities with the true likelihood of outcomes, ensuring that the confidence levels expressed by the model are accurate and reliable~\cite{wang2024, Joy2023}. A well-calibrated model produces probabilities that reflect real-world frequencies; for instance, if a model predicts an event with 70\% confidence, that event should occur approximately 70\% of the time. Calibration is particularly important in applications where decision-making relies on the trustworthiness of probabilistic outputs e.g., medical diagnosis. Poorly calibrated models can either overestimate or underestimate their confidence, leading to overconfidence or underconfidence in predictions, which can have significant consequences in critical domains~\cite{wang2024}. 

Temperature Scaling (TS)~\cite{Guo2017} is a post-hoc calibration method that is commonly employed for model calibration.
In TS, the calibrated probabilities for label $y$ given the input $x$ is defined as:
\begin{equation}
P(y|x) = \text{softmax}\left(\frac{z}{T}\right)    
\end{equation}
where $z$ are the logits--raw score outputs from the model that are transformed into probabilities, $T$ is the temperature parameter.

A higher value of $T$ smooths the probability distribution, reducing overconfidence, while a lower value of $T$ sharpens it. TS uses a single global $T$, optimized on a validation set, to improve calibration across all samples. However, this approach assumes that all samples require the same adjustment, which is often impractical~\cite{Joy2023}.

Adaptive Temperature Scaling (ATS) proposed in the study by Joy et al.~\cite{Joy2023} is an extension of traditional TS, designed to
dynamically adjust the temperature parameter for each individual sample or context. ATS introduces a sample-dependent temperature function $T(x)$ that adapts based on the characteristics of each input $x$. This approach improves the calibration of deep learning models, especially in scenarios with diverse data distributions.

A variant of ATS is proposed in the study by Balanya et al.~\cite{Balanya2024}, that combines two approaches: linear temperature scaling and entropy based temperature scaling. The combined model, denoted as $\text{HnLTS}$, defines the temperature function as
\begin{equation}
T_{\text{HnLTS}}(z) = \sigma_{\text{SP}}\big(w^L z + w^H \log \bar{H}(z)),
\end{equation}
where $w^L z$ represents a linear transformation of the logits $z$, capturing their direct influence on the temperature. The term $w^H \log \bar{H}(z)$ incorporates the logarithm of the normalized entropy $\bar{H}(z)$, reflecting the uncertainty in the predictions and enabling the model to adjust the temperature based on confidence levels. The function $\sigma_{\text{SP}}$, typically a softplus activation, ensures that the temperature remains positive. By integrating both the logits and the uncertainty, $\text{HnLTS}$ dynamically adapts the temperature for each sample, achieving robust and flexible calibration across diverse data distributions~\cite{Balanya2024}.

\section{Methods}
\label{sec:methods}

In this study, we use the ResNet-18 architecture to classify cleansing quality in \ac{CCE} images. For our experiments, we use the classical three-step pruning strategy of initial training, pruning, and fine-tuning~\cite{Han2015}, with 13 pruning steps.
The ResNet-18 model is configured for a 4-class classification problem, and includes a dropout rate of 0.4  before the final fully connected classification layer and  uses L1 norm to determine the importance of channels for pruning. For each pruning step, we prune 20\% of the channels of the convolutional layers of ResNet-18. We skip the first convolutional layer, as the first convolutional layer is critical for initial feature extraction, and pruning it can significantly degrade the model performance~\cite{Hoefler2021}.

We use the cross entropy loss function~\cite{goodfellow2016deep} and the Adam optimizer. To prevent overfitting we employ L2 regularization via the weight decay parameter and early stopping with a patience value of 5.

We employ stratified 10-fold cross validation, which maintains proportional representation of each cleansing category across training and validation sets. This approach minimizes overfitting and enhances the generalizability of the model. 

We apply various image-level augmentation techniques~\cite{domben2023} to enhance the robustness of our model. To account for symmetry and orientation variations, we use random horizontal flipping and random rotation. For adapting to different lighting conditions, we employ color jitter.

\subsection{Explainability}

In order to compare the explainability of the different pruned versions of the models, in line with~\cite{jacobgilpytorchcam} we employ the metrics Grad-CAM~\cite{Selvaraju2017}, Grad-CAM++~\cite{Chattopadhay2018}, Eigen-CAM~\cite{Bany2021}, Ablation-CAM~\cite{Desai2020}, and Random-CAM~\cite{jacobgilpytorchcam}.

Grad-CAM is a visualization technique that is used to explain deep learning models, especially \acp{CNN}. It shows the important regions in an input image that influence the model’s predictions by using gradients of the output class with respect to the last convolutional layer of the deep learning model~\cite{Selvaraju2017}.

Grad-CAM++ improves Grad-CAM by considering higher-order gradient information, making it more precise for identifying multiple important regions in a given image. In this manner, it enhances the localization of important features in an image and works better in cases where multiple objects are present~\cite{Chattopadhay2018}.

Eigen-CAM uses \ac{PCA} to generate more robust and sharp visual explanations. Unlike Grad-CAM, which relies on gradients, Eigen-CAM extracts the most dominant patterns from feature maps, making it relatively gradient-free and efficient visualization technique~\cite{Bany2021}.

Ablation-CAM is a gradient-free visualization technique used to generate \acp{CAM} for \acp{CNN}. Unlike Grad-CAM, Ablation-CAM removes certain neurons from the feature maps and observes their impact on the model’s output~\cite{Desai2020}. In other words, the importance of each feature map is measured by the drop in class score when it is removed.

Random-CAM is a baseline method that generates random importance weights instead of using gradients or ablation. It generates \acp{CAM} with random uniform values in the range [-1,1].

\begin{figure}[!ht]
\centering
\includegraphics[width = \textwidth]{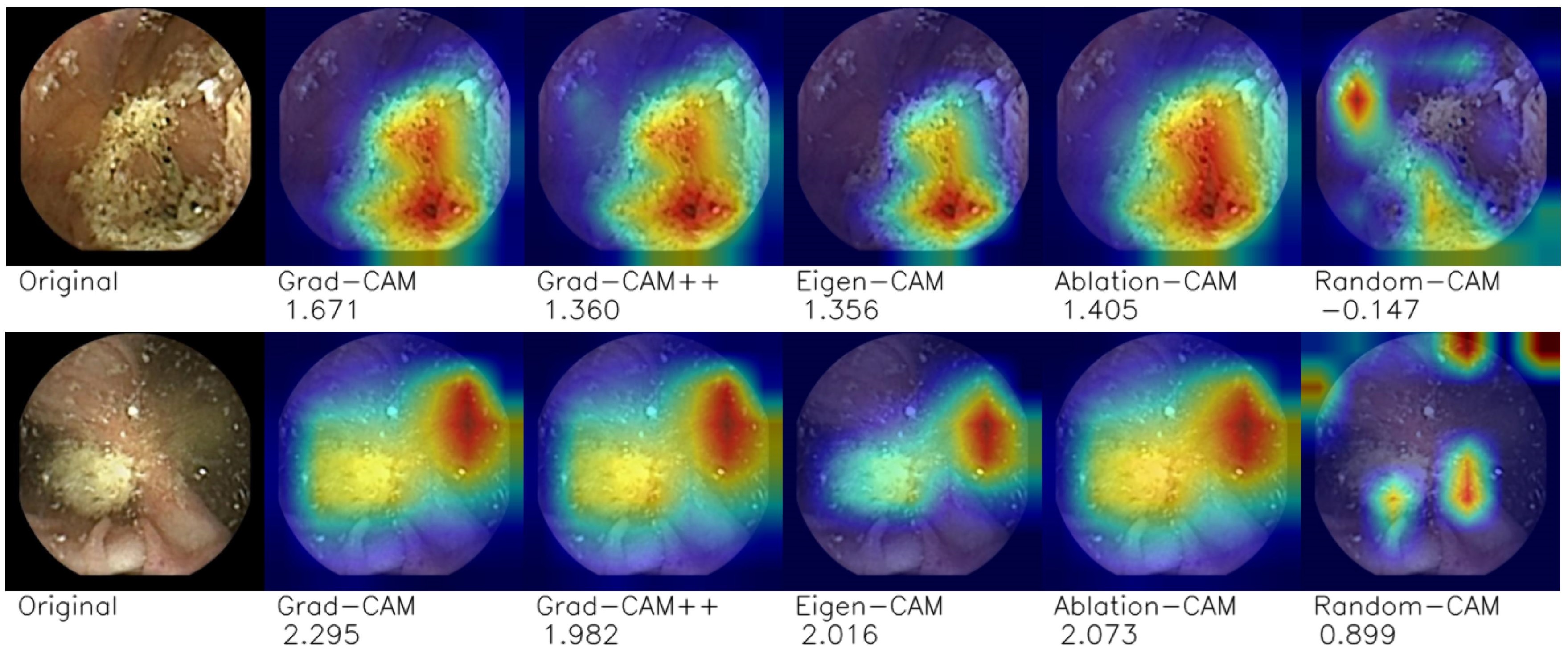}
\centering
\caption{Scores obtained for two images (titled Original) belonging to the ``Poor'' category of Leighton-Rex, and visualizations and scores of the different metrics.}
\label{fig:scores_example}
\end{figure}

In Figure~\ref{fig:scores_example}, we show the ROAD~\cite{ROAD2022} scores obtained by using Grad-CAM, Grad-CAM++, Eigen-CAM, Ablation-CAM, and Random-CAM metrics for the original model ( discussed later) for two example images belonging to the category: Poor. We can note that for the selected two examples, scores associated with Grad-CAM, Grad-CAM++, Eigen-CAM, Ablation-CAM can be different, for the original images in the first and second column, the Grad-CAM scores are highest. Furthermore, the scores from the metrics: Grad-CAM++, Eigen-CAM, and Ablation-CAM are higher than that of the scores from Random-CAM.

\subsection{Model Calibration}

Combined Temperature Scaling (HnLTS) model from the study~\cite{Balanya2024} that combines linear scaling of logits (LTS) with entropy-based scaling (HTS) is used for our experiments. 
The model calibration is done on the external dataset, by using 25 percent of the external dataset (99 images) for validation to minimize the cross-entropy loss, with early stopping (patience value of 5) and the rest of the 297 images are used for testing. The HTS consists of a fully connected layer that maps the deep learning model's features size to a hidden layer with $h$ neurons, followed by a ReLU activation  to introduce non-linearity. The output of the hidden layer is passed through another fully connected layer to produce a single scalar value per input sample, which is then activated using the Softplus function to ensure the output is strictly positive. For our experiments we used  three different values for the hidden units $h = 16, 32, 64$. 

\section{Results}
\label{sec:results}

In this section, we discuss the results obtained by using the strategy of training, pruning, and fine-tuning~\cite{Han2015} on the cleansing quality dataset as outlined in Section~\ref{sec:data}. 

\begin{figure}[!ht]
\centering
\includegraphics[width = 0.85\textwidth]{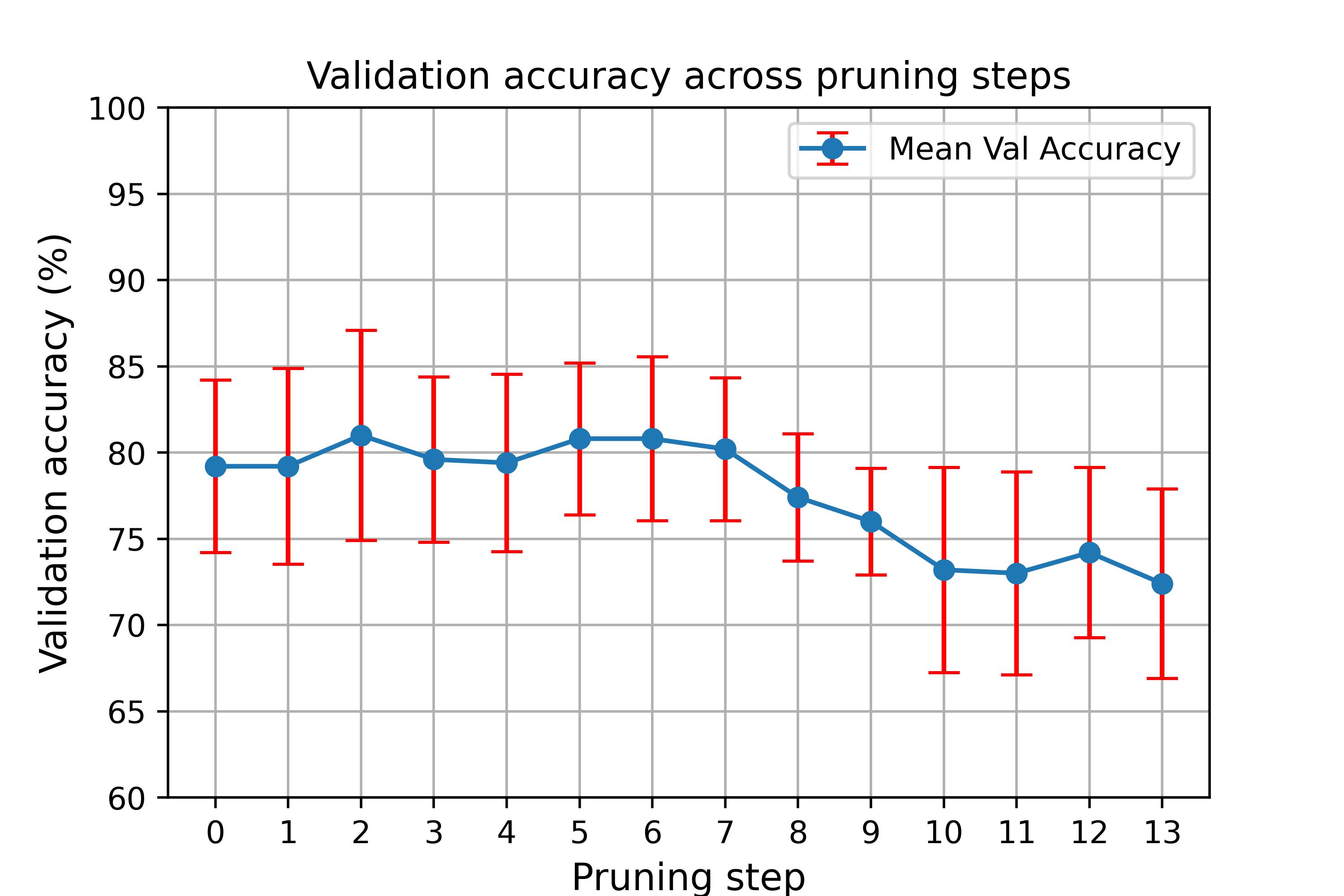}
\centering
\caption{Mean cross-validation accuracy and $\pm$ one standard deviation across 10 folds versus pruning steps.}
\label{fig:meanaccuracy_per_step}
\end{figure}

We use 13 pruning steps, where the 0th step is without any pruning, i.e., the original model. The mean accuracy associated with 10 folds across different pruning steps is shown in Figure~\ref{fig:meanaccuracy_per_step}. We use mean accuracy across 10 folds as measure of average performance of the model. We can observe that the mean accuracy at step 0 is 79.20\% and best accuracy is 84.00\%. For more details, please see Table~\ref{tab:results}. The figure shows that the mean accuracies for pruning step 2 to 7 are more than that mean accuracy at step 0. These results are in line with other studies~\cite{Hao2016, zhu2017, blalock2020} which indicate that pruning can initially improve the performance of deep learning models by reducing overfitting and enhancing generalization. From Table~\ref{tab:results}, we can note that the mean accuracy at step 7 is 80.20\% (best accuracy is 88.00\%). After step 7, the mean accuracy starts to decrease, however, the best accuracy at step 11 (84.00\%) is still comparable to that of step 0. For our experiments on explainability and model calibration, we used the model(s) with best accuracy from each step.

\begin{table}[h]
\caption{Mean accuracy for the 10 folds, best accuracy among the 10 folds, and sparsity of the best model for different pruning steps.}
\label{tab:results}
\begin{tabular}{|c|c|c|c|}
\hline
Pruning step & Mean accuracy (\%) & Best accuracy (\%) & Sparsity (\%) \\
\hline
0 & 79.20 & 84.00 & 0.00 \\
\hline
1 & 79.20 & 88.00 & 19.94 \\
\hline
2 & 81.00 & 90.00 & 35.93 \\
\hline
3 & 79.60 & 86.00 & 48.80 \\
\hline
4 & 79.40 & 88.00 & 58.96 \\
\hline
5 & 80.80 & 88.00 & 67.18 \\
\hline
6 & 80.80 & 88.00 & 73.79 \\
\hline
7 & 80.20 & 88.00 & 79.03 \\
\hline
8 & 77.40 & 84.00 & 83.16 \\
\hline
9 & 76.00 & 80.00 & 86.51 \\
\hline
10 & 73.20 & 80.00 & 89.27 \\
\hline
11 & 73.00 & 84.00 & 91.38 \\
\hline
12 & 74.20 & 82.00 & 93.09 \\
\hline
13 & 72.40 & 82.00 & 94.50 \\
\hline
\end{tabular}
\end{table}

\begin{figure}[!ht]
\centering
\includegraphics[width = 0.75\textwidth]{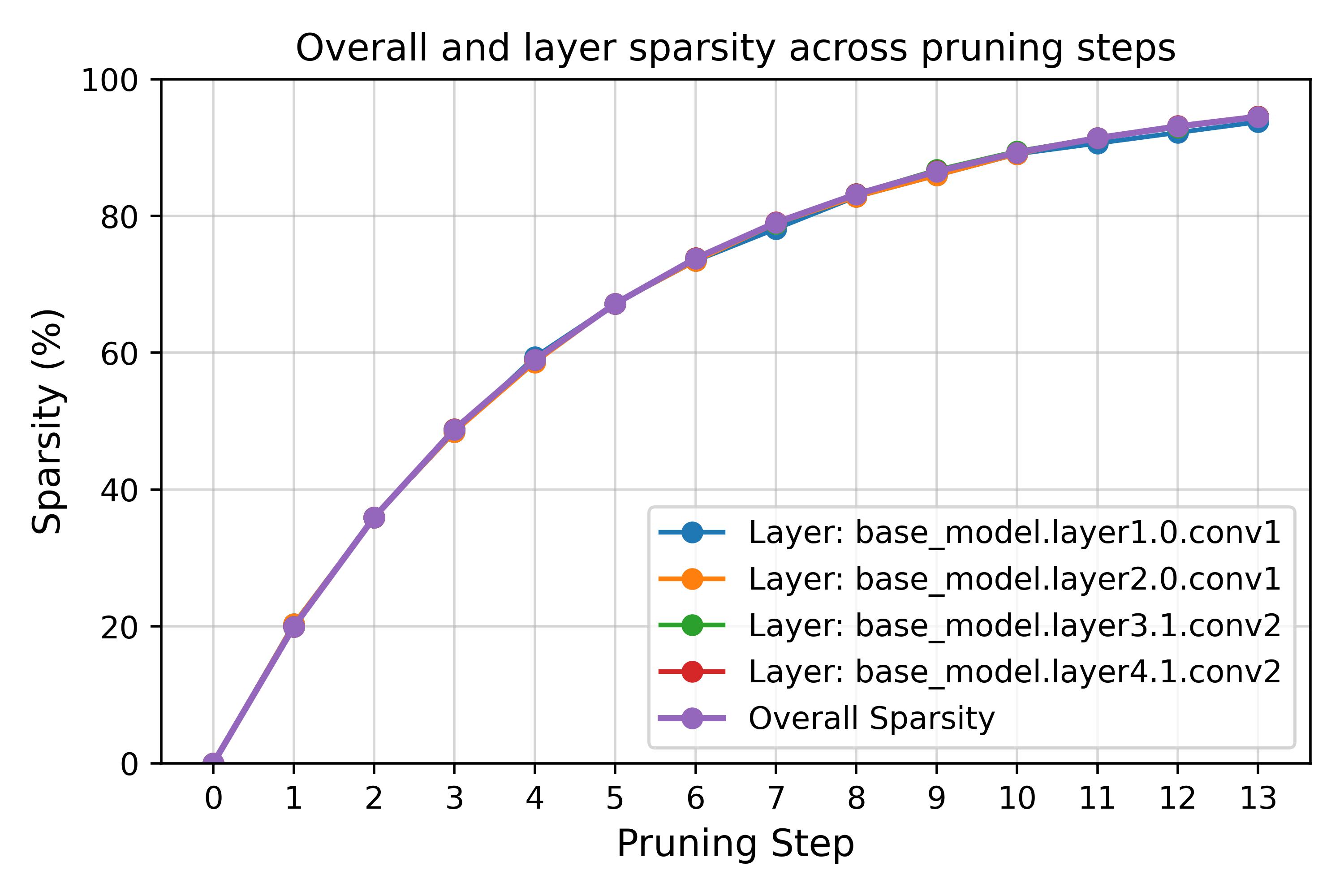}
\caption{Overall sparsity and layer sparsity for pruning steps.}
\label{fig:overall_sparsity_plots}
\end{figure}

We show the percentage of overall sparsity for each pruning step in Figure~\ref{fig:overall_sparsity_plots}. The sparsity calculation is performed at both the layer level and the overall model level by analyzing the proportion of zero-valued parameters to the total number of parameters in the weights. As the 0th step corresponds to the original model its overall sparsity is 0\%. When we compare it with Figure~\ref{fig:meanaccuracy_per_step}, we can see that at step 7 (at mean accuracy of  80.20\%), the sparsity is 79.03\%. 
We have used a pruning fraction of 0.2 (20\%) for multiple pruning steps, the sparsity does not increase linearly by 20\% at each step as the pruning fraction is applied to the remaining unpruned weights and not the original total number of weights. This is something that can observed in Figure~\ref{fig:overall_sparsity_plots}, where the overall sparsity is nearly 20\% for 0th step, is approx. 36\% for 1st step, and follows a similar trend for later steps. We also show (in Figure~\ref{fig:overall_sparsity_plots}) the sparsity percentages per layer for four selected convolutional layers of the model, the trend across the four layers is the same for majority of the pruning steps.

\subsection{Results for explainability}

In this section, we discuss the results associated with using the metrics Grad-CAM, Grad-CAM++, Eigen-CAM, Ablation-CAM, and Random-CAM for a subset of 64 images. The subset was selected by randomly sampling 16 images from each of the four cleansing categories. 

\begin{figure}[!ht]
\centering
\includegraphics[width = 0.75\textwidth]{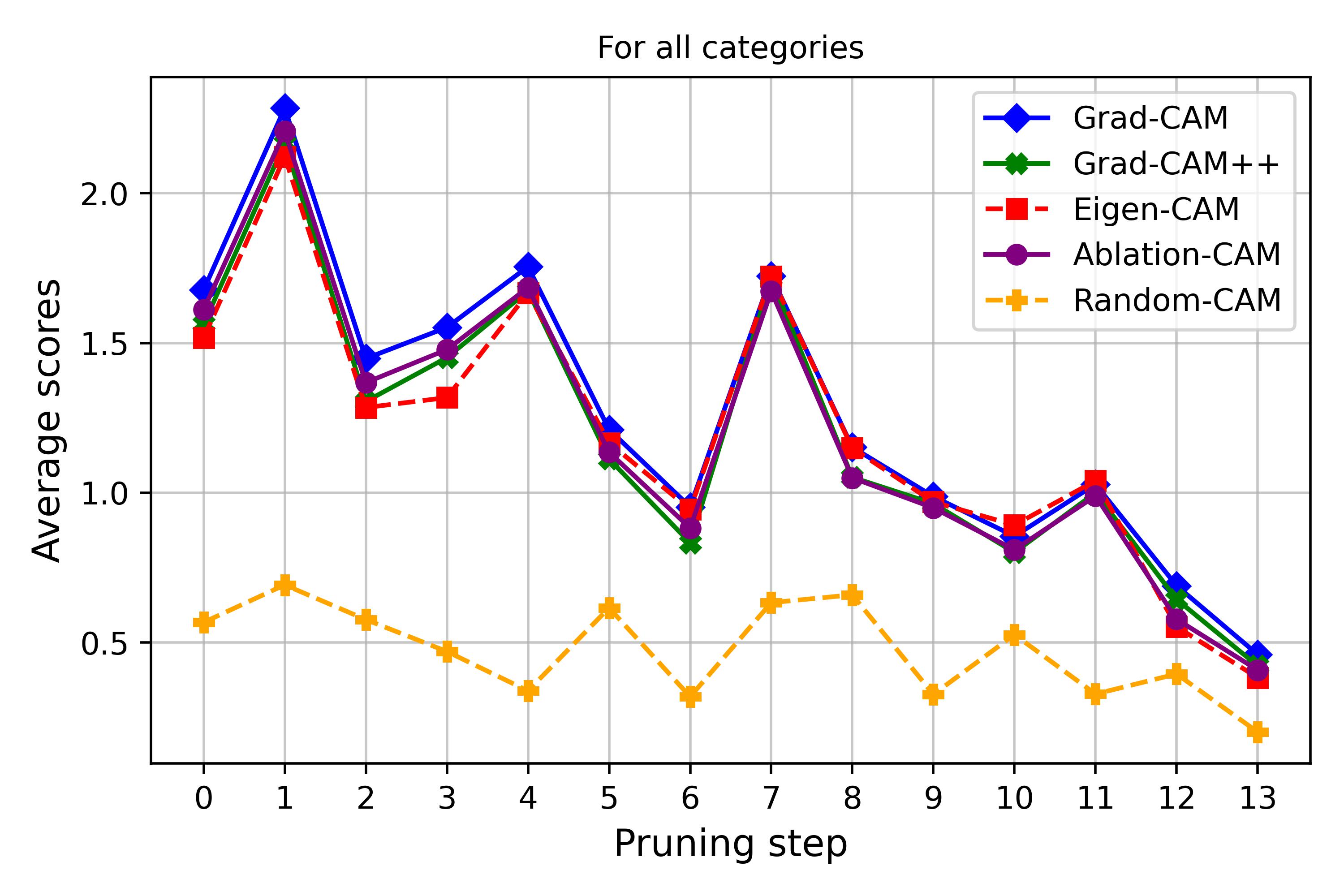}
\caption{Scores from the metrics for all images and categories selected for analysis.}
\label{fig:scores_all_image_categories}
\end{figure}

The results associated with score calculations are shown in Figure~\ref{fig:scores_all_image_categories}. For each pruning step, we plot the mean scores for the five metrics used in the paper. The mean scores from Random-CAM are lower than all other metrics. For a majority of the steps, the mean scores for Grad-CAM, Grad-CAM++, Eigen-CAM and Ablation-CAM are consistent with each other. As the pruning steps increase, i.e., as the model is pruned more,  the gap between the scores of Random-CAM and other genuine metrics starts to decrease.

\begin{figure}[!ht]
\centering
\includegraphics[width = 0.75\textwidth]{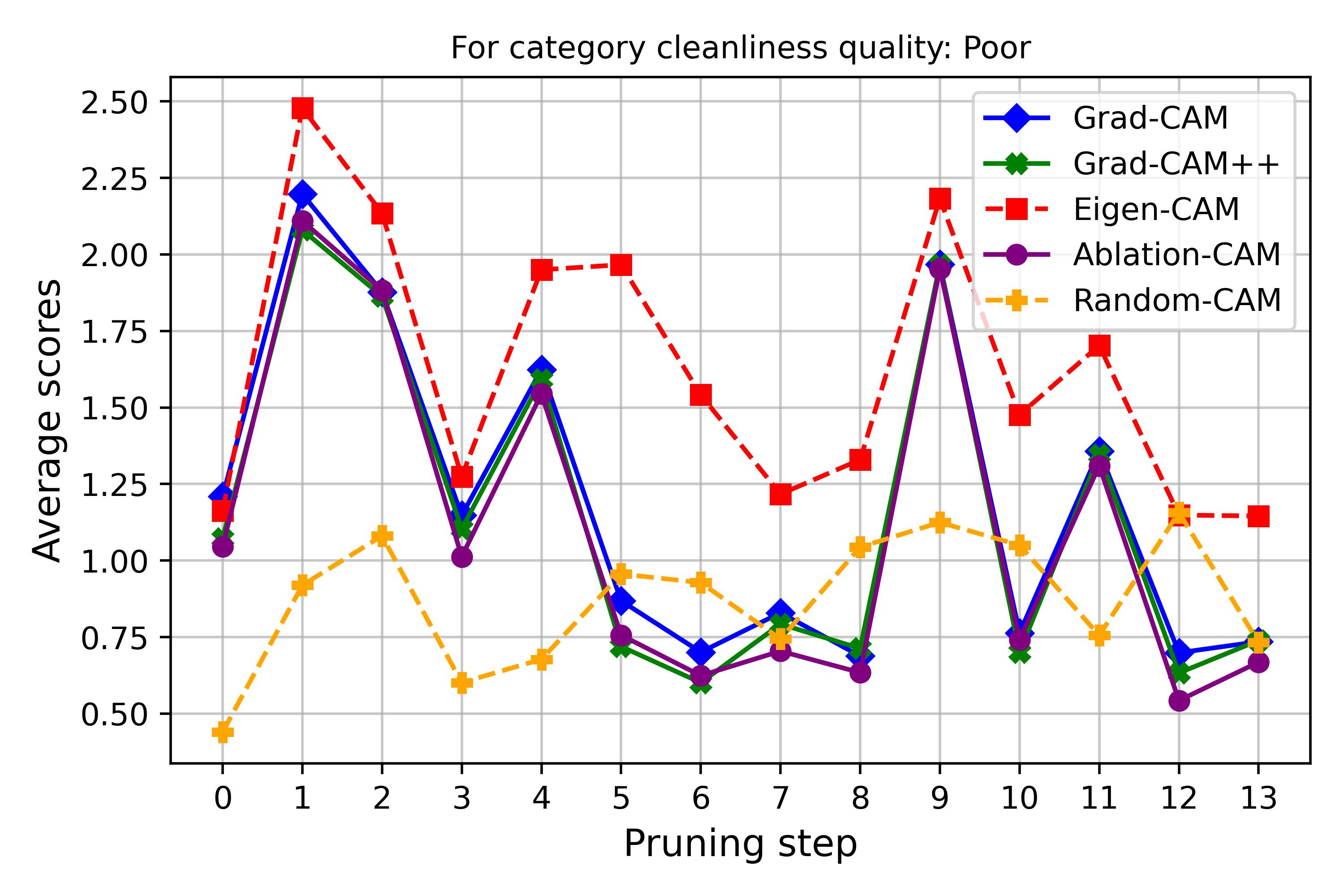}
\caption{Mean scores obtained by using the different metrics for images belonging to category: Poor.}
\label{fig:scores_category_0}
\end{figure}

\begin{figure}[!ht]
\centering
\includegraphics[width = 0.75\textwidth]{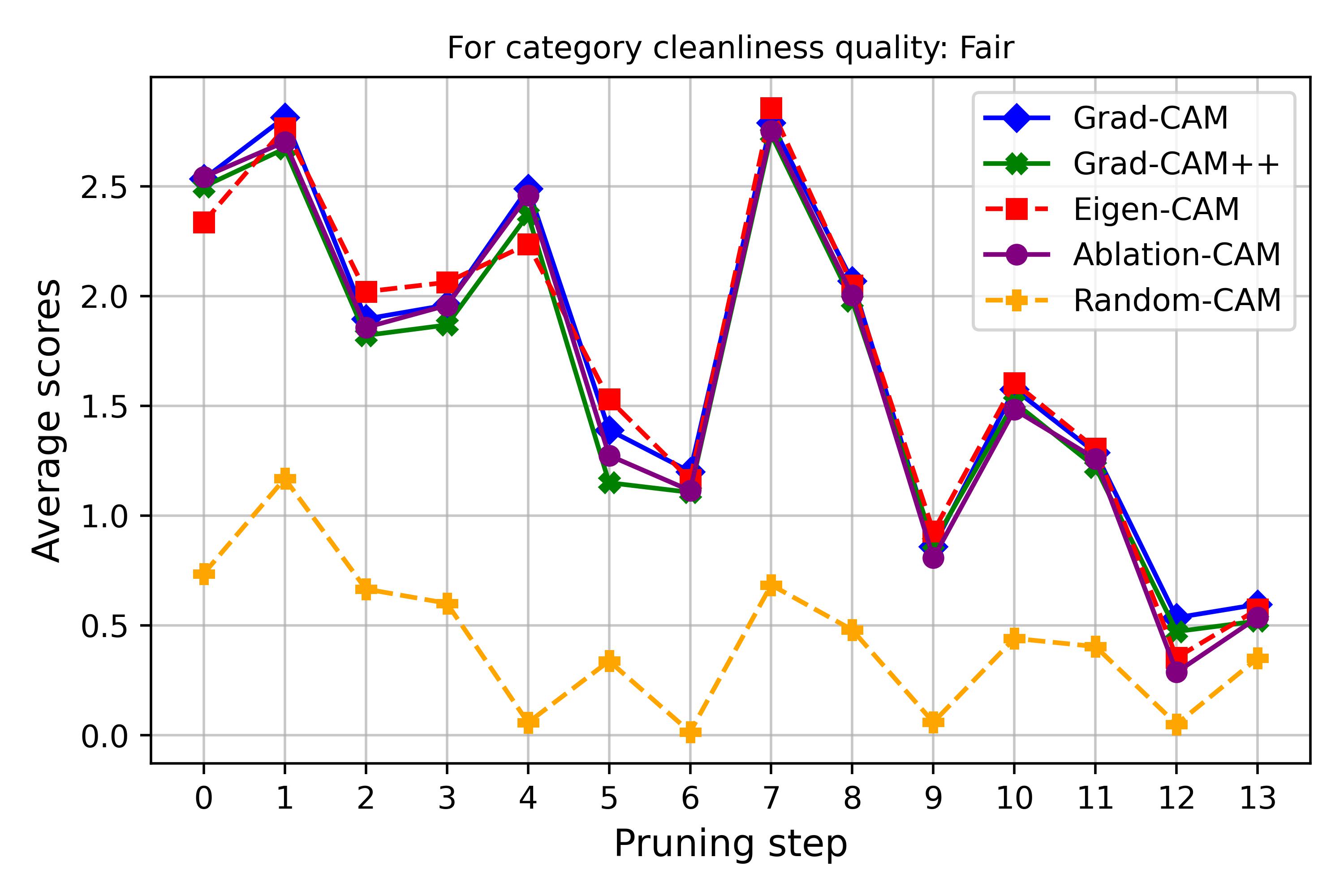}
\caption{Mean scores obtained by using the different metrics for images belonging to category: Fair}
\label{fig:scores_category_1}
\end{figure}

\begin{figure}[!ht]
\centering
\includegraphics[width = 0.75\textwidth]{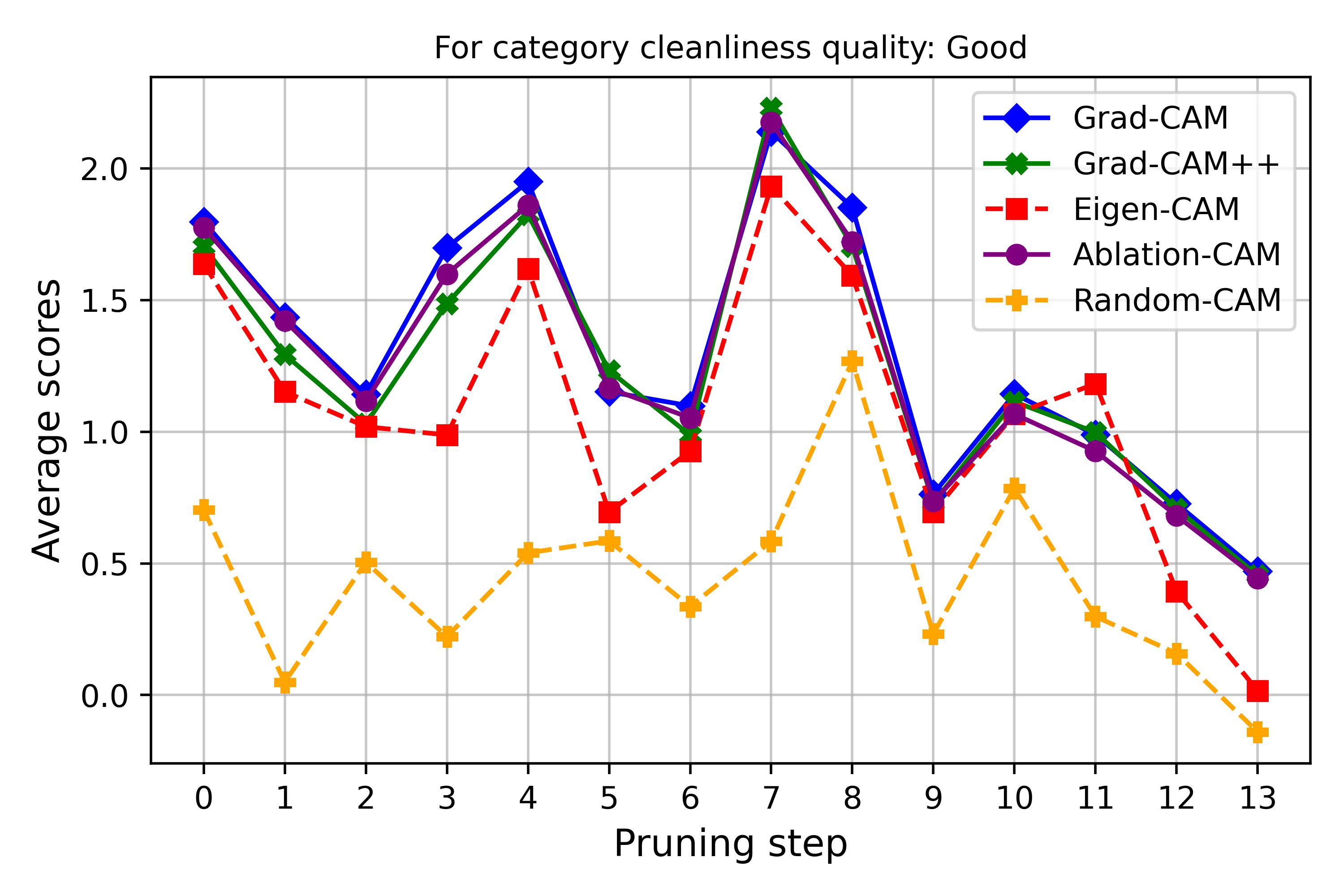}
\caption{Mean scores obtained by using the different metrics for images belonging to category: Good}
\label{fig:scores_category_2}
\end{figure}

\begin{figure}[!ht]
\centering
\includegraphics[width = 0.75\textwidth]{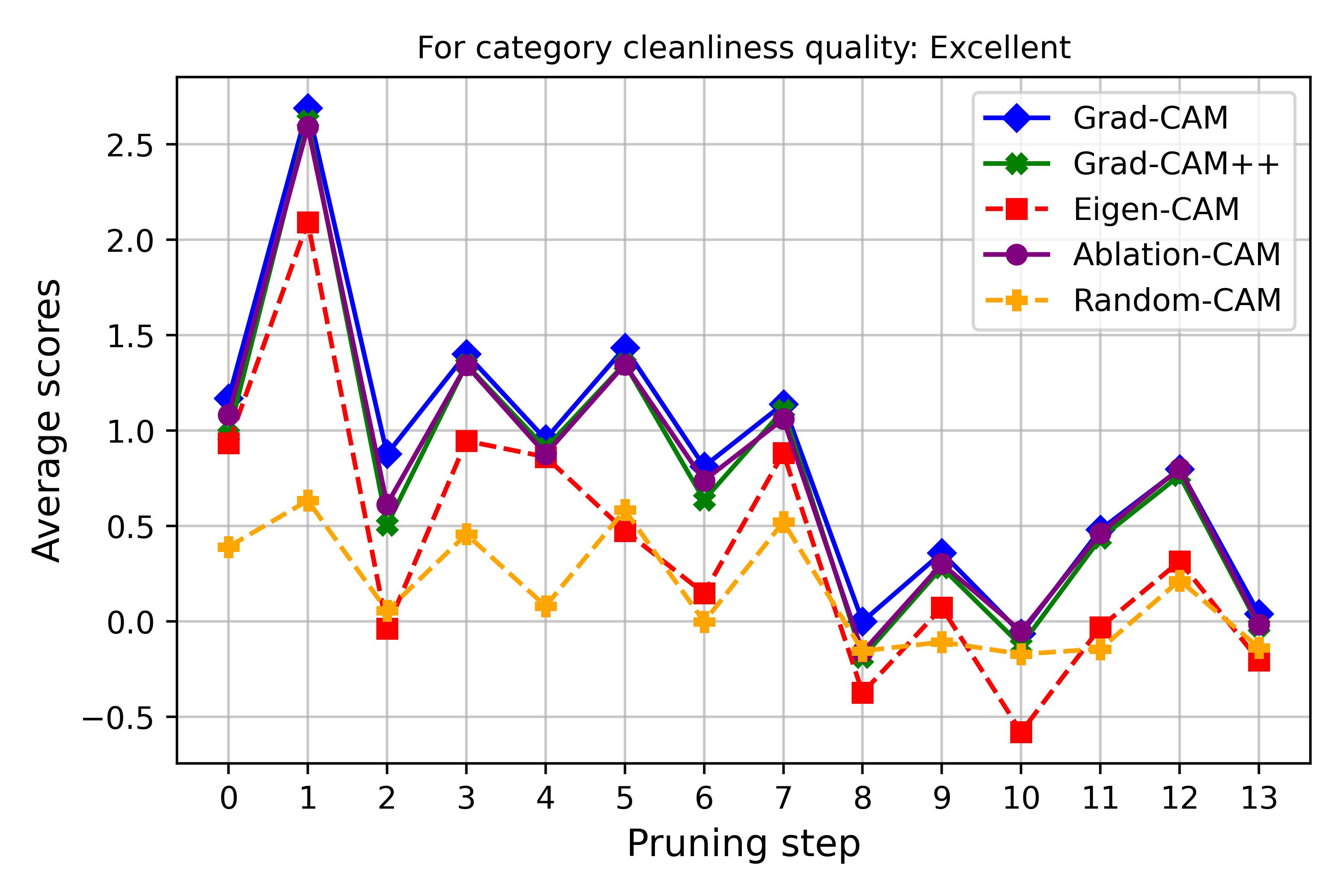}
\caption{Mean scores obtained by using the different metrics for images belonging to category: Excellent}
\label{fig:scores_category_3}
\end{figure}

In order to verify if the mean scores of the five metrics are consistent across the four Leighton-Rex cleansing categories, we also plot the scores associated with the 16 images of each category in Figures~\ref{fig:scores_category_0},~\ref{fig:scores_category_1},~\ref{fig:scores_category_2}, and~\ref{fig:scores_category_3}.  These figures reveal notable variations in the mean scores for the different metrics across the four categories, highlighting the influence of label categories on the observed trends. The detailed discussion below further elaborates on these variations and their implications--the relationship between cleansing categories and metric scores.

The cleansing quality categories ``Poor'' and ``Fair'' are of particular importance as these are crucial regarding whether the \ac{CCE} can be classified as complete. It is important to note depending on the Clinicians, Poor and Fair categories can be considered inadequate and Good and Excellent categories can be adequate for the completion of the the \ac{CCE} investigation.

Figure~\ref{fig:scores_category_0} shows that the gap between the mean score from Random-CAM and other genuine metrics is less for ``Poor'' images. In fact, for pruning steps 5, 6, 7, 8, 10, 12, and 13, it is either comparable or more than the mean scores of other metrics.

From the Figures~\ref{fig:scores_category_1},~\ref{fig:scores_category_2} associated with categories ``Fair'' and ``Good'', we observe that these results are similar to trends observed in Figure~\ref{fig:scores_all_image_categories} where the mean scores of metrics Grad-CAM, Grad-CAM++, Eigen-CAM, and  Ablation-CAM are consistently better than that of mean scores from Random-CAM. 

In the ``Excellent'' category as shown in Figure~\ref{fig:scores_category_0}, for a majority of steps, the mean scores for Random CAM are less than that of the scores from other metrics with the exception of steps 8 and 13 where they are quite comparable to the mean scores of metrics Grad-CAM, Grad-CAM++, Eigen-CAM, and  Ablation-CAM.  

\begin{figure}[!ht]
\centering
\includegraphics[width = \textwidth]{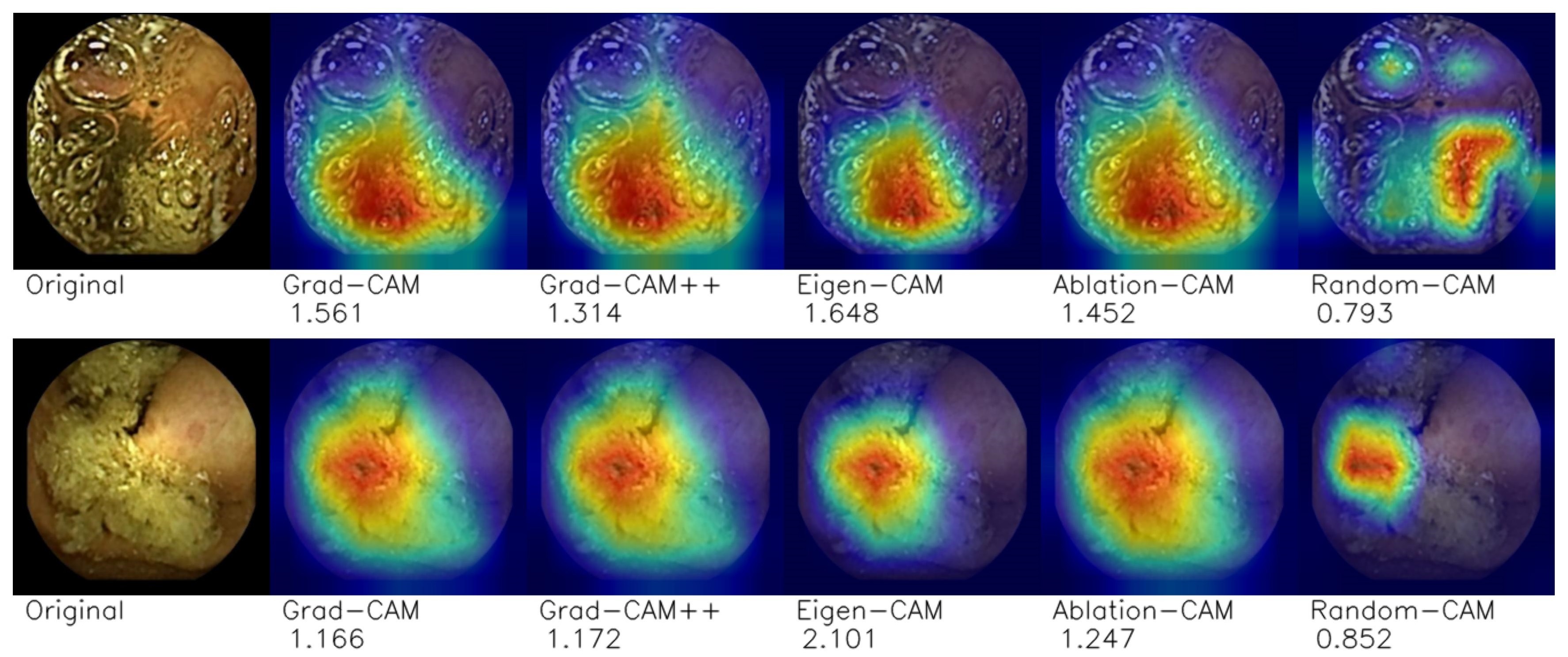}
\caption{Different ROAD metric scores for two example images belonging to category: Poor using the model from pruning step 7.}
\label{fig:some_examples_of_poor_cat_pos}
\end{figure}

\begin{figure}[!ht]
\centering
\includegraphics[width = \textwidth]{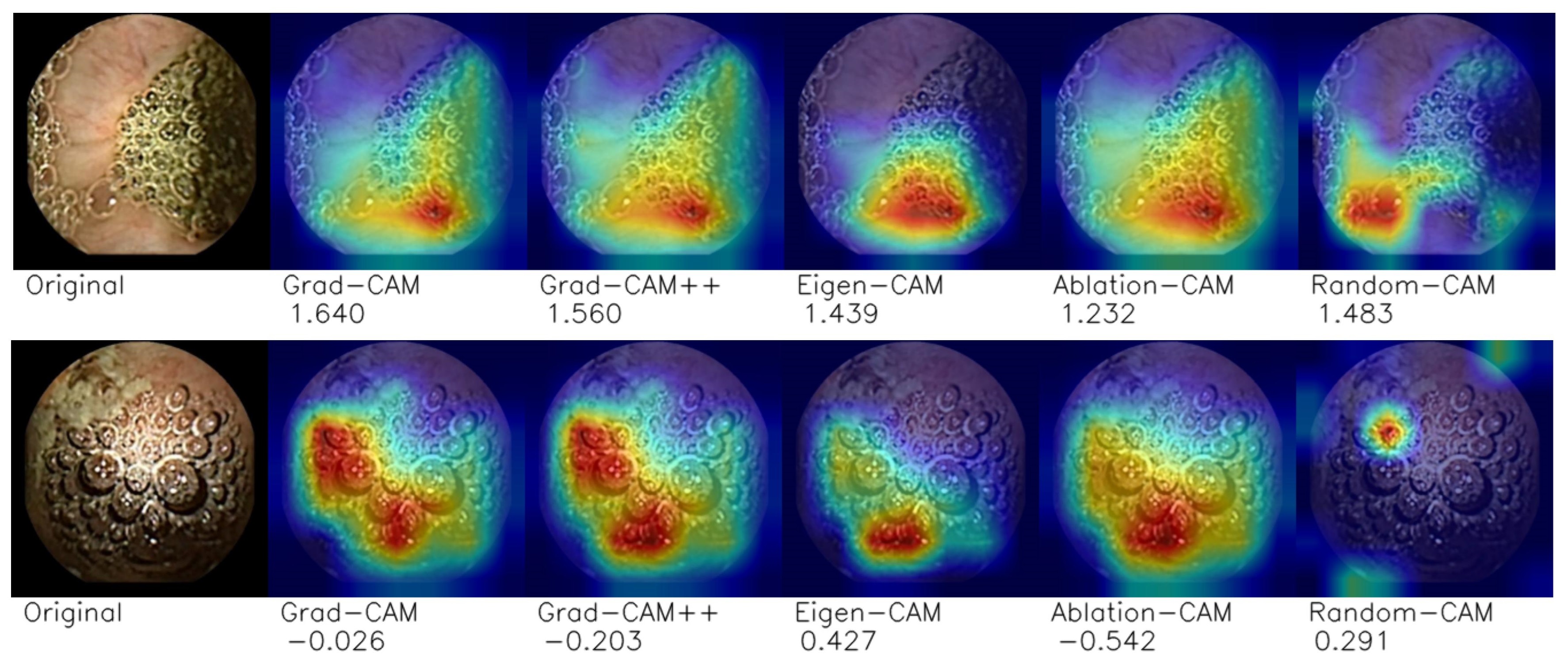}
\caption{Different ROAD metric scores for two example images belonging to category: Poor using the model from pruning step 7.}
\label{fig:some_examples_of_poor_cat_neg}
\end{figure}

\begin{figure}[!ht]
\centering
\includegraphics[width = \textwidth]{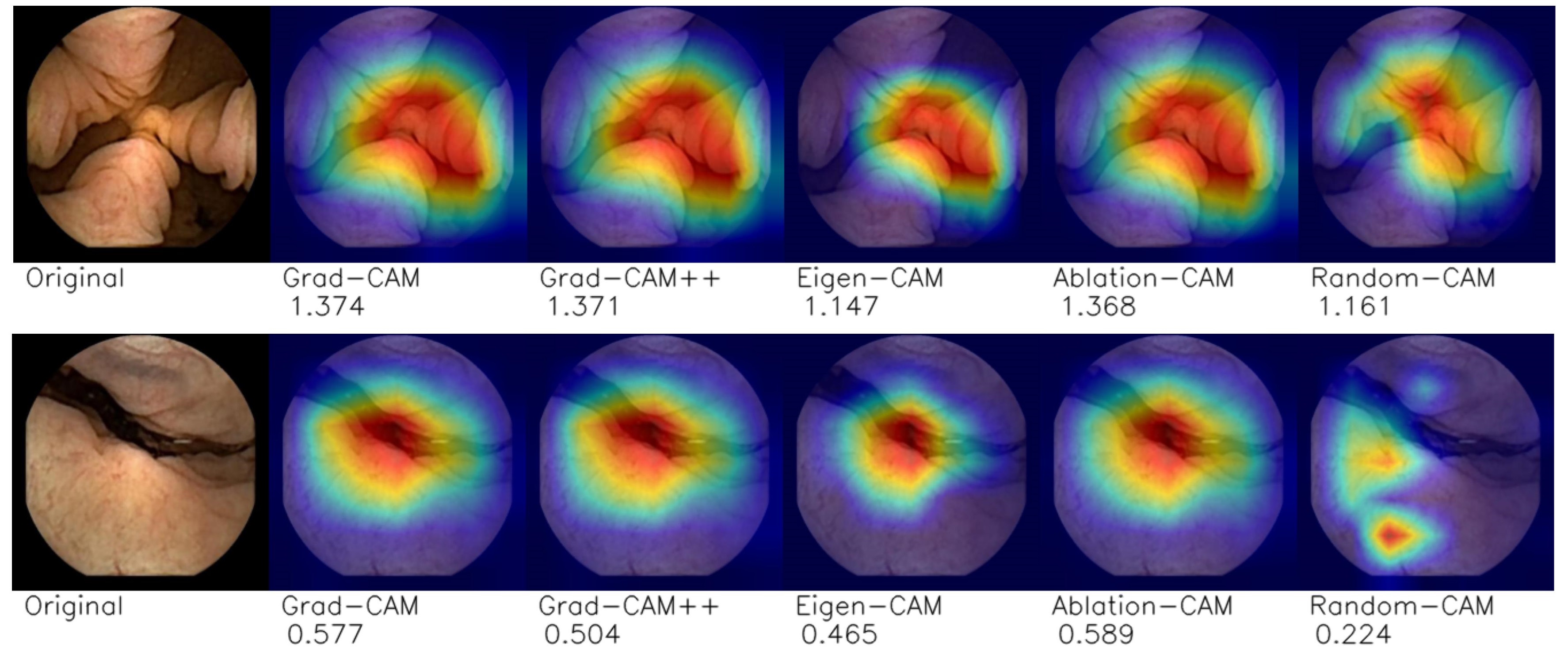}
\caption{Different ROAD metric scores for two example images belonging to category: Excellent using the model from pruning step 7.}
\label{fig:some_examples_of_excellent_cat}
\end{figure}

In Figure~\ref{fig:some_examples_of_poor_cat_pos}, we show two examples of images from the category ``Poor'' and their associated scores for the metrics: Grad-CAM, Grad-CAM++, Eigen-CAM, Ablation-CAM, and Random-CAM. We can note that the scores for Random-CAM are less than that of the scores from other metrics that indicates a good classification for these type of images.
Next, we show two examples of images for the category ``Poor'' in Figure~\ref{fig:some_examples_of_poor_cat_neg}, where we can note that the scores from Random-CAM are comparable or in some cases more than that of the scores obtained from other metrics. 
Figure~\ref{fig:some_examples_of_excellent_cat} shows images belonging to the category: Excellent and their associated visualizations using the ROAD framework~\cite{ROAD2022} with metrics scores. We observe that for the original image in the first row, the score from Random-CAM has a large positive value and is in similar range to that of the other genuine metrics. These observations underscore the limitations of relying on such evaluation metrics—depending on the specific category (e.g., cleansing category ``Poor'' in our case) and task, the scores may lack the comprehensiveness needed for informed decision-making, thereby warranting careful consideration and sometimes categorical breakdown of the analysis.

\subsection{Results for calibration}

\begin{figure}[!ht]
\centering
\includegraphics[width = 0.75\textwidth]{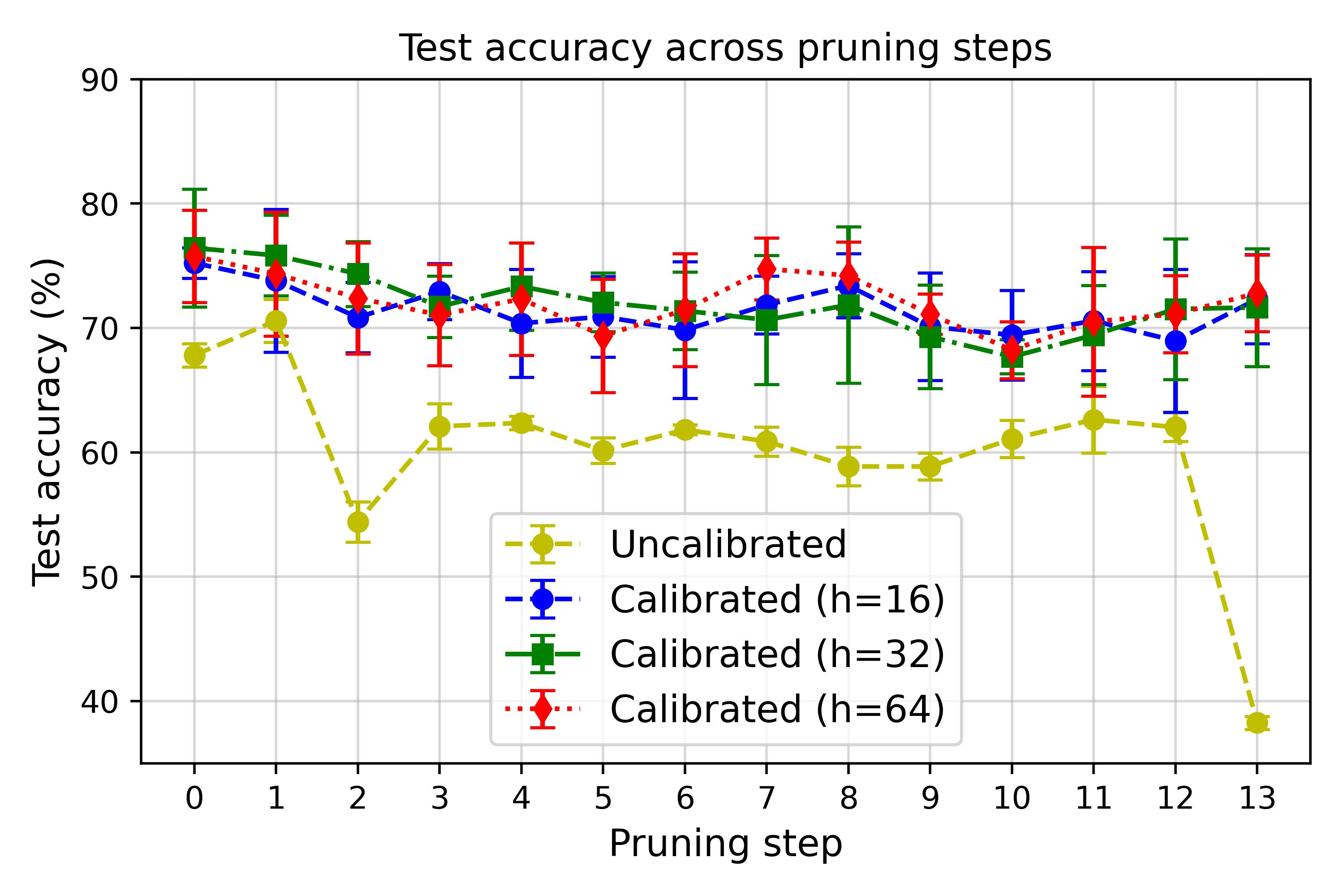}
\caption{Test accuracies for calibrated and uncalibrated models over pruning steps with hidden unit sizes 16, 32, and 64.}
\label{fig:calibrated_test_accuracies}
\end{figure}

\begin{figure}[!ht]
\centering
\includegraphics[width = \textwidth]{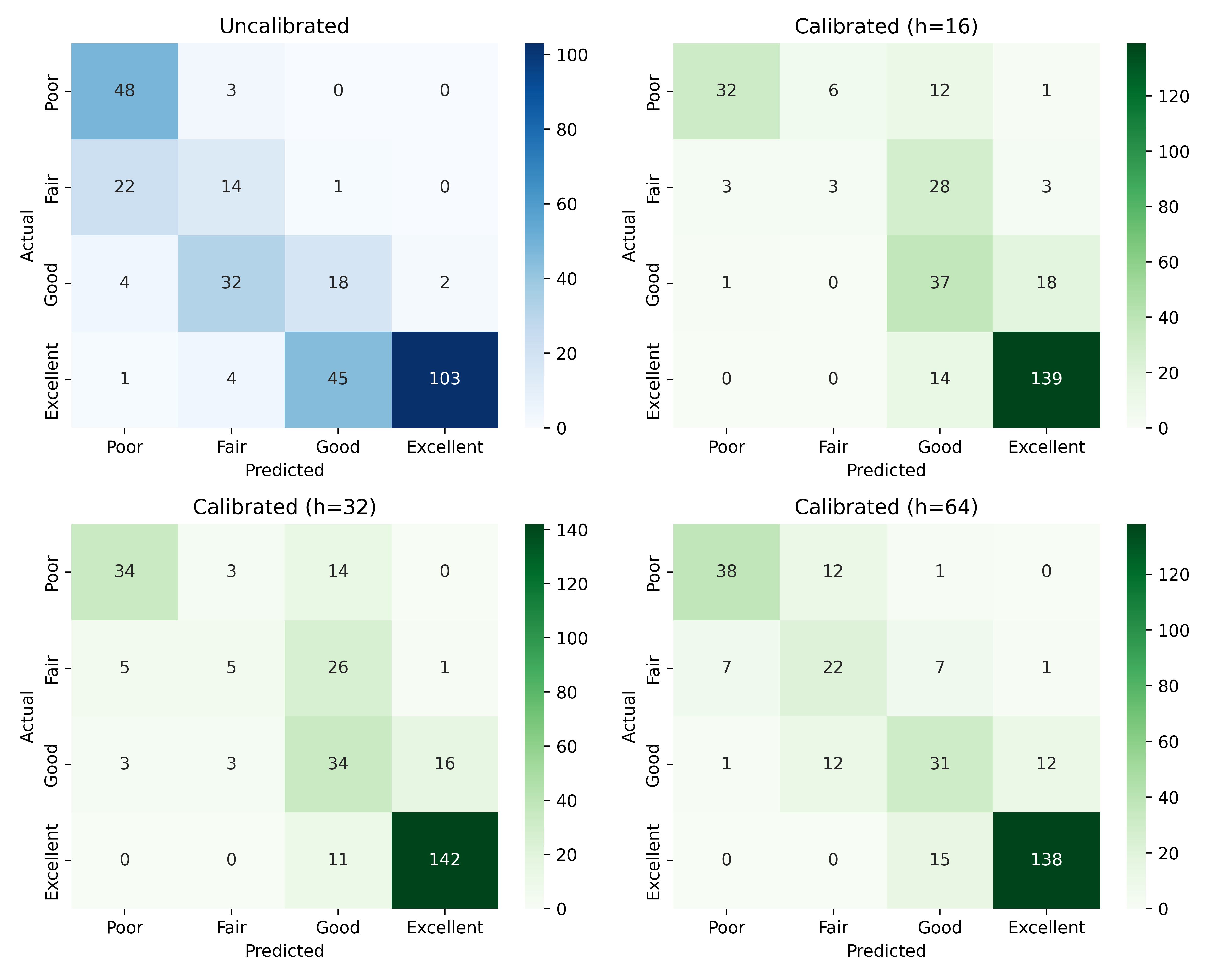}
\caption{Confusion matrices for test accuracies for Uncalibrated and Calibrated models over pruning steps with hidden unit sizes 16, 32, and 64.}
\label{fig:confusion_matrices}
\end{figure}

In order to provide a robust evaluation of the model's performance after using the Combined Temperature Scaling (HnLTS) method for model's calibration, we performed the experiments 5 times with the results reported as mean test accuracy accompanied by standard deviation. 
The model calibration was performed on the external dataset. 

Figure~\ref{fig:calibrated_test_accuracies} shows test accuracies for the external dataset and pruned model(s) across different pruning steps. Wee see that the calibrated models seem to perform better than that of uncalibrated models for all pruned steps. For instance, at step 7 the uncalibrated mean test accuracy is 61.61\%, for hidden unit sizes, h = 16, 32, and 64 the calibrated models give test accuracies of 71.04\%, 72.39\%, and 77.10\%, respectively. To have detailed view on the influence of calibration on the classifications across individual categories of data, Figure~\ref{fig:confusion_matrices} shows the confusion matrices for uncalibrated model and calibrated models using different number of hidden units for one run of the experiments. We note that calibration of the model leads to improvement in the classification associated with the categories: ``Good'' and ``Excellent'' at the expense of decrease in classification associated with the categories: ``Poor'' and ``Fair'' for h = 16, 32. As the external data has higher number of samples associated with the category: ``Excellent'', calibration leads to an improvement in the overall performance of the model(s). We also note that for hidden unit, h= 64, there is an improvement in the three classification categories namely ``Fair'',``Good'' and ``Excellent'-- indicating a better calibrated model.

\section{Discussion}

This study demonstrates the potential of deep learning models to automatically assess cleansing quality in \ac{CCE} images. The iterative pruning strategy proves effective in reducing model complexity while maintaining high accuracy. For instance, we achieved a maximum cross-validation accuracy of 88\% with 79\% sparsity at pruning step 7 improving efficiency from 84\% corresponding to the original model. This highlights the feasibility of deploying such models on resource-constrained devices, which is critical for real-world clinical applications.

Our study also underscores the challenges associated with evaluating cleansing quality. The explainability metrics Grad-CAM, Grad-CAM++, Eigen-CAM, and  Ablation-CAM for the ROAD framework can prove valuable for checking the correspondence between the performance and its explainability. However, as our results suggest, the consistency for the different metrics can vary and hence careful measures (such as detailed analysis of different categories) are needed to ensure the model's predictions align with clinical expectations. 

Another important aspect to note is although the explainability metrics for the ROAD framework are post-hoc and hence considered to be computationally efficient, they may not fully capture the underlying  mechanisms driving the model's decisions. In addition, in order to calculate the scores we still need to run the inference on a subset of images and use a single target layer i.e., last convolutional layer of a model, and both factors can influence the fidelity and robustness of the explanations. The choice of the subset and the specific layer can introduce biases or limit the generalizability of the insights, especially in complex models where decisions may be influenced by multiple layers or features that are not captured in the selected configuration. As a result, while ROAD framework~\cite{ROAD2022} provides valuable insights for interpretability, its dependence on these design choices must be carefully considered when drawing conclusions from its outputs.

Overall, this study provides a strong foundation for leveraging deep learning in \ac{CCE} cleansing image analysis, with implications for improving the efficiency and accuracy of clinical workflows.

Future work should focus on addressing the limitations of the dataset, such as incorporating more diverse samples and refining labeling processes to reduce subjectivity. Additionally, exploring advanced pruning techniques such as sparse training~\cite{Hoefler2021, frankle2019, evci2021} i.e., introducing sparsity in the model's weights, activations, or gradients during training.

The improvement in performance underscores the effectiveness of the HnLTS calibration method in enhancing the model's predictive reliability, particularly for pruned architectures derived from limited data. However, this improvement is not uniformly distributed across all categories. Calibration of the model leads to better classification accuracy for the three categories ``Fair'', ``Good'', ``Excellent'' but this comes at the expense of a decrease in classification accuracy for the categories ``Poor''. Overall it leads to a better calibrated model based on a hidden unit size of 64. This study also highlights the importance of performing model calibration with careful consideration, ensuring its impact on different classification categories is thoroughly analyzed.

\section{Conclusion}\label{sec13}

This study demonstrates the potential of deep learning models to automatically assess cleansing quality in \acf{CCE} images. By employing an iterative pruning strategy, we effectively reduced model complexity while maintaining high accuracy, achieving a maximum cross-validation accuracy of 88\% with 79\% sparsity at pruning step 7—an improvement in accuracy compared to the original model's 84\%.
The study also underscores the challenges of evaluating cleansing quality and the importance of explainability in deep learning models. Metrics such as Grad-CAM, Grad-CAM++, Eigen-CAM, and Ablation-CAM within the ROAD framework provide valuable insights into model interpretability. However, our findings reveal variability in the consistency of these metrics, necessitating careful analysis to ensure alignment between model predictions and clinical expectations. While the ROAD framework offers valuable interpretability tools, its design dependencies must be carefully considered.
The study further highlights the effectiveness of the HnLTS calibration method in improving predictive reliability, particularly for pruned architectures derived from limited data. Calibration enhanced classification accuracy for the categories "Fair," "Good," and "Excellent," though it leads to a decrease in accuracy for the "Poor" category. This trade-off underscores the importance of performing model calibration with careful consideration of its impact on different classification categories.

\backmatter

\section*{Declarations}


This research is part of AICE project (number 101057400) funded by the European Union, and it is part-funded by the United Kingdom government. Views and opinions expressed are however those of the author(s) only and do not necessarily reflect those of the European Union or the European Commission. Neither the European Union nor the European Commission can be held responsible for them.

The data used in this article is part of the CFC2015 trial's outcome, which belongs to the “Odense University Hospital (OUH)”. The request to access the data, or part of it should be made to Prof. Gunnar Baatrup. See GIT-LINK for Python code.

\begin{figure}[H]
    \centering
   \includegraphics[height=1.1cm]{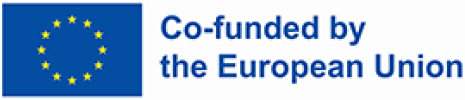}
    \includegraphics[height=1.1cm]{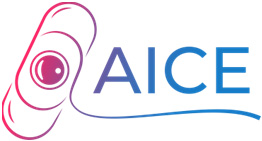}
    \label{fig:funding}
\end{figure}

\bibliography{diagnostics}

\end{document}